\documentclass[11pt,a4paper]{article}
\usepackage[hyperref]{acl2021}
\usepackage{times}
\usepackage{latexsym}

\usepackage{balance}
\usepackage[linesnumbered,ruled,vlined,algo2e]{algorithm2e}
\usepackage{bm}
\usepackage{amssymb}
\usepackage{amsmath}
\usepackage{url}
\usepackage{makecell}
\usepackage{multirow}
\usepackage{subfigure}
\usepackage{graphicx} 
\usepackage{framed}
\usepackage{color}
\usepackage{colortbl}
\usepackage{textcomp}
\usepackage{CJKutf8}
\usepackage{booktabs}
\usepackage{microtype}
\usepackage{arydshln}
\usepackage{xspace}
\usepackage{amsmath}
\usepackage{enumitem}
\usepackage{amssymb}
\usepackage{esvect}
\usepackage{color}
\usepackage{caption}
\usepackage{braket}

\usepackage{tikz}

\renewcommand\vec[1]{\overrightarrow{#1}}
\newcommand\cev[1]{\overleftarrow{#1}}

\usepackage{color}
\usepackage{colortbl}
\definecolor{g1}{RGB}{232,232,232}
\definecolor{g2}{RGB}{207,207,207}
\definecolor{g3}{RGB}{181,181,181}
\definecolor{g4}{RGB}{156,156,156}

\definecolor{b1}{RGB}{217,217,254}
\definecolor{b2}{RGB}{198,198,253}
\definecolor{b3}{RGB}{180,180,252}
\definecolor{b4}{RGB}{162,162,252}
\definecolor{b5}{RGB}{136,136,255}
\definecolor{r1}{RGB}{254,236,236}
\definecolor{r2}{RGB}{254,217,217}
\definecolor{r3}{RGB}{253,198,198}
\definecolor{r4}{RGB}{253,180,180}
\definecolor{r5}{RGB}{252,128,127}

\definecolor{electricviolet}{rgb}{0.56, 0.0, 1.0}

\definecolor{takeaway}{RGB}{209,226,206}

\definecolor{forestgreen}{rgb}{0.0, 0.50, 0.0}
\definecolor{goldenbrown}{rgb}{0.6, 0.4, 0.08}

\DeclareMathOperator*{\argmax}{arg\,max}

\aclfinalcopy 
\def\aclpaperid{3448} 

\title{Rejuvenating Low-Frequency Words: \\ Making the Most of Parallel Data in Non-Autoregressive Translation}

\author{
Liang Ding\thanks{~~~Liang Ding and Longyue Wang contributed equally to this work. Work was done when Liang Ding and Xuebo Liu were interning at Tencent AI Lab.}\\
The University of Sydney\\
\normalsize \textsf{ldin3097@sydnye.edu.au}\\
\And
Longyue Wang$^*$\\
Tencent AI Lab\\
\normalsize \textsf{vinnylywang@tencent.com}\\
\And
Xuebo Liu\\
University of Macau\\
\normalsize \textsf{nlp2ct.xuebo@gmail.com}\\
\AND
Derek F. Wong\\
University of Macau\\
\normalsize \textsf{derekfw@um.edu.com}\\
\And
Dacheng Tao\\
JD Explore Academy, JD.com\\
\normalsize \textsf{dacheng.tao@gmail.com}\\
\And
Zhaopeng Tu\\
Tencent AI Lab\\
\normalsize \textsf{zptu@tencent.com}\\
}

\date{}

\begin{document}

\maketitle
\begin{abstract}
Knowledge distillation (KD) is commonly used to construct synthetic data for training non-autoregressive translation (NAT) models. However, there exists a discrepancy on low-frequency words between the distilled and the original data, leading to more errors on predicting low-frequency words. To alleviate the problem, we directly expose the raw data into NAT by leveraging pretraining. By analyzing directed alignments, we found that KD makes low-frequency source words aligned with targets more deterministically but fails to align sufficient low-frequency words from target to source. Accordingly, we propose reverse KD to rejuvenate more alignments for low-frequency target words. To make the most of authentic and synthetic data, we combine these complementary approaches as a new training strategy for further boosting NAT performance. We conduct experiments on five translation benchmarks over two advanced architectures. Results demonstrate that the proposed approach can significantly and universally improve translation quality by reducing translation errors on low-frequency words. Encouragingly, our approach achieves 28.2 and 33.9 BLEU points on the WMT14 English-German and WMT16 Romanian-English datasets, respectively. Our code, data, and trained models are available at \url{https://github.com/alphadl/RLFW-NAT}. 
\end{abstract}

\section{Introduction}
Recent years have seen a surge of interest in non-autoregressive translation~(\citealp[NAT,][]{NAT}), which can improve the decoding efficiency by predicting all tokens independently and simultaneously.
The {\em non-autoregressive factorization} breaks conditional dependencies among output tokens,
which prevents a model from properly capturing the highly multimodal distribution of target translations. 
As a result, the translation quality of NAT models often lags behind that of autoregressive translation~(\citealp[AT,][]{transformer}) models.
To balance the trade-off between decoding speed and translation quality, knowledge distillation (KD) is widely used to construct a new training data for NAT models~\cite{NAT}. Specifically, target sentences in the distilled training data are generated by an AT teacher, which makes NAT easily acquire more deterministic knowledge and achieve significant improvement~\cite{zhou2019understanding}.
 
Previous studies have shown that distillation may lose some important information in the original training data, leading to more errors on predicting low-frequency words. To alleviate this problem, \citet{Ding2020UnderstandingAI} proposed to augment NAT models the ability to learn lost knowledge from the original data. 
However, their approach relies on external resources (e.g. word alignment) and human-crafted priors, which limits the applicability of the method to a broader range of tasks and languages.
Accordingly, we turn to directly expose the raw data into NAT by leveraging pretraining without intensive modification to model architectures (\S\ref{subsec:2.2}). Furthermore, we analyze bilingual links in the distilled data from two alignment directions (i.e. source-to-target and target-to-source). We found that KD makes low-frequency source words aligned with targets more deterministically but fails to align low-frequency words from target to source due to information loss. Inspired by this finding, we propose reverse KD to recall more alignments for low-frequency target words (\S\ref{subsec:2.3}). We then concatenate two kinds of distilled data to maintain advantages of deterministic knowledge and low-frequency information. To make the most of authentic and synthetic data, we combine three complementary approaches (i.e. raw pretraining, bidirectional distillation training and KD finetuning) as a new training strategy for further boosting NAT performance (\S\ref{subsec:2.4}).

\begin{figure*}[t]
    \centering
    \subfigure[Traditional Training]{
    \includegraphics[height=0.14\textheight]{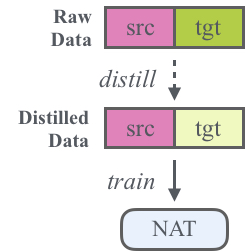}}
    \hspace{0.03\textwidth}
    \subfigure[Raw Pretraining]{
    \includegraphics[height=0.14\textheight]{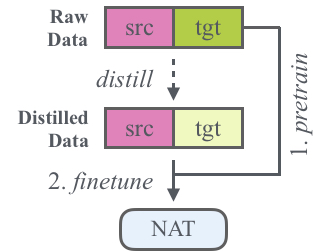}}
    \hspace{0.03\textwidth}
    \subfigure[Bidirectional Distillation Training]{
    \includegraphics[height=0.14\textheight]{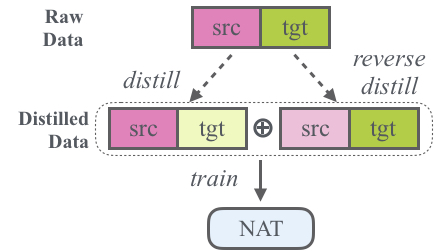}}
    \caption{An illustration of different strategies for training NAT models. ``distill'' and ``reverse distill'' indicate sequence-level knowledge distillation with forward and backward AT teachers, respectively. The data block in transparent color means source- or target-side data are synthetically generated. Best view in color.}
    \label{fig:framework}
\end{figure*}

We validated our approach on five translation benchmarks (WMT14 En-De, WMT16 Ro-En, WMT17 Zh-En, WAT17 Ja-En and WMT19 En-De)
over two advanced architectures (\citealp[Mask Predict,][]{ghazvininejad2019mask}; \citealp[Levenshtein Transformer,][]{gu2019levenshtein}).
Experimental results show that the proposed method consistently improve translation performance over the standard NAT models across languages and advanced NAT architectures. Extensive analyses confirm that the performance improvement indeed comes from the better lexical translation accuracy especially on low-frequency tokens.
\paragraph{Contributions} 
{Our main contributions} are:
\begin{itemize}[leftmargin=12pt]
\item {We show the effectiveness of rejuvenating low-frequency information by pretraining NAT models from raw data.}
\item {We provide a quantitative analysis of bilingual links to demonstrate the necessity to improve low-frequency alignment by leveraging both KD and reverse KD.}
\item {We introduce a simple and effective training recipe to accomplish this goal, which is robustly applicable to several model structures and language pairs.}
\end{itemize}

\section{Rejuvenating Low-Frequency Words}
\subsection{Preliminaries}
\label{subsec:2.1}
\paragraph{Non-Autoregressive Translation} Given a source sentence $\bf x$, an AT model generates each target word ${\bf y}_t$ conditioned on previously generated ones ${\bf y}_{<t}$, leading to high latency on the decoding stage. In contrast, NAT models break this {\em autoregressive factorization} by producing target words in parallel. Accordingly, the probability of generating $\bf y$ is computed as:
\begin{equation}
    p({\bf y}|{\bf x})
    =\prod_{t=1}^{T}p({\bf y}_t|{\bf x}; \theta)
\end{equation}
where $T$ is the length of the target sequence, and it is usually predicted by a separate conditional distribution. The parameters $\theta$ are trained to maximize the likelihood of a set of training examples according to $\mathcal{L}(\theta) = \argmax_{\theta} \log p({\bf y}|{\bf x}; \theta)$. Typically, most NAT models are implemented upon the framework of Transformer~\cite{transformer}.

\paragraph{Knowledge Distillation} \citet{NAT} pointed out that NAT models suffer from the {\em multimodality problem}, where the conditional independence assumption prevents a model from properly capturing the highly multimodal distribution of target translations. Thus, the sequence-level knowledge distillation is introduced to reduce the modes of training data by replacing their original target-side samples with sentences generated by an AT teacher~\cite{NAT,zhou2019understanding,ren2020astudy}. Formally, the original parallel data {\em Raw} and the distilled data {\em $\vec{\text{KD}}$} can be defined as follows:
\begin{align}
    \text{Raw} &= \{(\mathbf{x}_i, \mathbf{y}_i)\}^N_{i=1}\\
    \vec{\text{KD}} &= \{(\mathbf{x}_i, f_{s \mapsto t}(\mathbf{x}_i))|\mathbf{x}_i \in \text{Raw}_\text{s}\}^N_{i=1}
    \label{eq:fkd}
\end{align}
where $f_{s \mapsto t}$ represents an AT-based translation model trained on $\text{Raw}$ data for translating text from the source to the target language. $N$ is the total number of sentence pairs in training data. As shown in Figure~\ref{fig:framework} (a), well-performed NAT models are generally trained on $\vec{\text{KD}}$ data instead of $\text{Raw}$.

\subsection{Pretraining with Raw Data}
\label{subsec:2.2}
\paragraph{Motivation}

\newcite{gao2018representation} showed that more than 90\% of words are lower than 10e-4 frequency in WMT14 En-De dataset. This {\em token imbalance problem} biases translation models towards over-fitting to frequent observations while neglecting those low-frequency observations~\cite{gong2018frage,nguyen-chiang-2018-improving,gu2020token}. Thus, the AT teacher $f_{s \mapsto t}$ tends to generate more high-frequency tokens and less low-frequency tokens during constructing distilled data $\vec{\text{KD}}$. 

On the one hand, KD can reduce the modes in training data (i.e. multiple lexical choices for a source word), which lowers the intrinsic uncertainty~\citep{ott2018analyzing} and learning difficulty for NAT~\citep{zhou2019understanding, ren2020astudy}, making it easily acquire more deterministic knowledge. On the other hand, KD aggravates the imbalance of high-frequency and low-frequency words in training data and lost some important information originated in raw data. \citet{Ding2020UnderstandingAI} revealed the side effect of distilled training data, which cause lexical choice errors for low-frequency words in NAT models. Accordingly, they introduced an extra bilingual data-dependent prior objective to augments NAT models the ability to learn the lost knowledge from raw data. 
We use their findings as our departure point, but rejuvenate low-frequency words in a more simple and direct way: directly exposing raw data into NAT via pretraining. 

\begin{table}[t]
\centering
    \begin{tabular}{lcccccc}
    \toprule
    \multirow{2}{*}{\bf Data} & \multicolumn{3}{c}{$\mathbf{s}\mapsto\mathbf{t}$ LFW Links}    & \multicolumn{3}{c}{$\mathbf{t}\mapsto\mathbf{s}$ LFW Links}\\
    \cmidrule(lr){2-4}
    \cmidrule(lr){5-7}
    &\bf R &\bf P  &\bf F1 &\bf R &\bf P  &\bf F1\\
    \midrule
    \bf Raw &66.4 &81.9 &73.3 &72.3 &80.6 & 76.2\\
    \bf $\vec{\text{KD}}$ &\bf 73.4 &\bf 89.2 &\bf 80.5 &69.9 &79.1 &74.2 \\
    \bf $\cev{\text{KD}}$ &61.2 &79.4 &69.1 &\bf 82.9 &\bf 83.1 &\bf 83.0\\
    \bottomrule
    \end{tabular}
    \caption{Evaluation on aligned links between source- and target-side low-frequency words (LFW). A directed line indicates aligning bilingual words from the source to the target side ($\mathbf{s}\mapsto\mathbf{t}$) or in an opposite way ($\mathbf{t}\mapsto\mathbf{s}$). R, P and F1 are recall, precision and F1-score.}
    \label{tab:analysis-values}
\end{table}

\paragraph{Our Approach} 

Many studies have shown that pretraining could transfer the knowledge and data distribution, especially for rare categories, hence improving the model robustness~\cite{hendrycks2019using,mathis2021pretraining}. Here we want to transfer the distribution of lost information, e.g. low-frequency words.
As illustrated in Figure~\ref{fig:framework}(b), we propose to first pretrain NAT models on $\text{Raw}$ data and then continuously train them on $\vec{\text{KD}}$ data. The raw data maintain the original distribution especially on low-frequency words. Although it is difficult for NAT to learn high-mode data, the pretraining can acquire general knowledge from authentic data, which may help {\em better} and {\em faster} learning further tasks. 
Thus, we early stop pretraining when the model can achieve 90\% of the best performance of raw data in terms of BLEU score~\cite{platanios2019competence}\footnote{{In preliminary experiments, we tried another simple strategy: early-stop at fixed step according to the size of training data (e.g. training 70K En-De and early stop at 20K / 30K / 40K, respectively). We found that both strategies achieve similar performance.}}. In order to keep the merits of low-modes, we further train the pretrained model on distilled data $\vec{\text{KD}}$. As it is easy for NAT to learn deterministic knowledge, we finetune the model for the rest steps. For fair comparison, the total training steps of the proposed method are same as the traditional one. In general, we expect that this training recipe can provide a good trade-off between raw and distilled data (i.e. high-modes and complete vs. low-modes and incomplete).

\begin{CJK}{UTF8}{gbsn}
\begin{table}[t]
\centering
\setlength{\tabcolsep}{2pt}
    \begin{tabular}{ll}
    \toprule
    \bf Data & \bf Sentence \\
    \midrule
    \bf Raw$_{\textbf{\tiny S}}$ & {\pdfliteral{2 Tr 0.2857 w}\textcolor{red}{海克曼}\pdfliteral{0 Tr}}\par~~和 {\pdfliteral{2 Tr 0.2857 w}\textcolor{blue}{奥德海姆}\pdfliteral{0 Tr}}\par~~提出 ... 模型\\
    \bf Raw$_{\textbf{\tiny T}}$ & \textcolor{goldenbrown}{\bf{Hackman}} and \textcolor{forestgreen}{\bf{Oldham}} propose ... model\\
    \midrule
    \bf $\vec{\text{KD}}_{\textbf{\tiny T}}$ & \textcolor{red}{\em Heckman} and \textcolor{blue}{\em Oddheim} propose ... model\\
    \bf $\cev{\text{KD}}_{\textbf{\tiny S}}$ & {\pdfliteral{1 0 0.3333 1 0 0 cm}\textcolor{goldenbrown}{哈克曼}\pdfliteral{1 0 -0.3333 1 0 0 cm}}\par~~和 {\pdfliteral{1 0 0.3333 1 0 0 cm}\textcolor{forestgreen}{奥尔德姆}\pdfliteral{1 0 -0.3333 1 0 0 cm}}\par~~提出 ... 模式 \\
    \bottomrule
    \end{tabular}
    \caption{An example in different kinds of data. ``Raw'' means the original data while ``$\vec{\text{KD}}$'' and ``$\cev{\text{KD}}$'' indicate syntactic data distilled by KD and reverse KD, respectively. The subscript ``S'' or ``T'' is short for source- or target-side. The low-frequency words are highlighted with colors and italics are incorrect translations.}
    \label{tab:analysis-example}
\end{table}
\end{CJK}

\subsection{Bidirectional Distillation Training}
\label{subsec:2.3}

\paragraph{Analyzing Bilingual Links in Data}
KD simplifies the training data by replacing low-frequency target words with high-frequency ones~\cite{zhou2019understanding}. This is able to facilitate easier aligning source words to target ones, resulting in high bilingual coverage~\cite{jiao2020data}. Due to the information loss, we argue that KD makes low-frequency target words have fewer opportunities to align with source ones. 
To verify this, we propose a method to quantitatively analyze bilingual links from two directions, where low-frequency words are aligned from source to target ($\text{s}\mapsto\text{t}$) or in an opposite direction ($\text{t}\mapsto\text{s}$). 

The method can be applied to different types of data. Here we take $\text{s}\mapsto\text{t}$ links in Raw data as an example to illustrate the algorithm. 
Given the WMT14 En-De parallel corpus, we employ an unsupervised word alignment method\footnote{The FastAlign~\citep{dyer2013simple} was employed to build word alignments for the training datasets.}~\cite{och2003systematic} to produce a word alignment, and then we extract aligned links whose source words are low-frequency (called $\text{s}\mapsto\text{t}$ LFW Links). Second, we randomly select a number of samples from the parallel corpus. For better comparison, the subset should contains the same $i$ in Equation (2) as that of other type of datasets (e.g. $i$ in Equation (3) for $\vec{\text{KD}}$). Finally, we calculate recall, precision, F1 scores based on low-frequency bilingual links for the subset. Recall (R) represents how many low-frequency source words can be aligned to targets. Precision (P) means how many aligned low-frequency links are correct according to human evaluation. F1 is the harmonic mean between precision and recall. Similarly, we can analyze $\text{t}\mapsto\text{s}$ LFW Links by considering low-frequency targets. 

Table~\ref{tab:analysis-values} shows the results on low-frequency links. Compared with Raw, $\vec{\text{KD}}$ can recall more $\text{s}\mapsto\text{t}$ LFW links (73.4 vs. 66.4) with more accurate alignment (89.2 vs. 73.3). This demonstrates the effectiveness of KD for NAT models from the bilingual alignment perspective. However, in the $\text{t}\mapsto\text{s}$ direction, there are fewer LFW links (69.9 vs. 72.3) with worse alignment quality (79.1 vs. 80.6) in $\vec{\text{KD}}$ than those in Raw. This confirms our claim that KD harms NAT models due to the loss of low-frequency target words. Inspired by these findings, it is natural to assume that {\em reverse KD} exhibits complementary properties. Accordingly, we conduct the same analysis method on $\cev{\text{KD}}$ data, and found better $\text{t}\mapsto\text{s}$ links but worse $\text{s}\mapsto\text{t}$ links compared with Raw. 
\begin{CJK}{UTF8}{gbsn}
Take the Zh-En sentence pair in Table~\ref{tab:analysis-example} for example, $\vec{\text{KD}}$ retains the source side low-frequency Chinese words ``海克曼'' (Raw$_\text{S}$) but generates the high-frequency English words ``Heckman'' instead of the golden ``Hackman'' ($\vec{\text{KD}}_{\textbf{\tiny T}}$). On the other hand, $\cev{\text{KD}}$ preserves the low-frequency English words ``Hackman'' (Raw$_\text{T}$) but produces the high-frequency Chinese words ``哈克曼'' ($\cev{\text{KD}}_{\textbf{\tiny S}}$). 
\end{CJK}

\paragraph{Our Approach}
Based on analysis results, we propose to train NAT models on bidirectional distillation by concatenating two kinds of distilled data.
The reverse distillation is to replace the source sentences in the original training data with synthetic ones generated by a backward AT teacher.\footnote{This is different from back-translation~\cite{edunov2018understanding}, which is an alternative to leverage monolingual data.} According to Equation~\ref{eq:fkd}, $\cev{\text{KD}}$ can be formulated as:
\begin{align}
    \cev{\text{KD}} &= \{(\mathbf{y}_i, f_{t \mapsto s}(\mathbf{y}_i))|\mathbf{y}_i \in \text{Raw}_\text{t}\}^N_{i=1}
\end{align}
where $f_{t \mapsto s}$ represents an AT-based translation model trained on $\text{Raw}$ data for translating text from the target to the source language.

Figure~\ref{fig:framework}(c) illustrates the training strategy. First, we employ both $f_{s \mapsto t}$ and $f_{t \mapsto s}$ AT models to generate $\vec{\text{KD}}$ and $\cev{\text{KD}}$ data, respectively. Considering complementarity of two distilled data, we combine $\vec{\text{KD}}$ and $\cev{\text{KD}}$ as a new training data for training NAT models. We expect that 1) distilled data can maintain advantages of low-modes; 2) bidirectinoal distillation can recall more LFW links on two directions with better alignment quality, leading to the overall improvements. 
Besides, \citet{nguyen2020data} claimed that combining different distilled data (generated by various models trained with different seeds) improves data diversification for NMT, and we leave this for future work.

\subsection{Combining Both of Them: Low-Frequency Rejuvenation (LFR)}
\label{subsec:2.4}
We have proposed two parallel approaches to rejuvenate low-frequency knowledge from authentic (\S\ref{subsec:2.2}) and synthetic (\S\ref{subsec:2.3}) data, respectively. Intuitively, we combine both of them to further improve the model performance. 

From data view, two presented training strategies are: $\text{Raw}\rightarrow\vec{\text{KD}}$ (Raw Pretraining) and $\vec{\text{KD}}+\cev{\text{KD}}$ (Bidirectional Distillation Training). Considering the effectiveness of pretraining~\cite{mathis2021pretraining} and clean finetuning~\cite{wu2019exploiting}, we introduce a combined pipeline: $\text{Raw}\rightarrow\vec{\text{KD}}+\cev{\text{KD}}\rightarrow\vec{\text{KD}}$ as out best training strategy. There are many possible ways to implement the general idea of combining two approaches. The aim of this paper is not to explore the whole space but simply to show that one fairly straightforward implementation works well and the idea is reasonable. Nonetheless, we compare possible strategies of combination two approaches as well as demonstrate their complementarity in \S\ref{sec:analysis}. While in main experiments (in \S\ref{subsec:3.2}), we valid the combination strategy, namely {\em Low-Frequency Rejuvenation} (LFR).

\begin{table*}[t]
    \centering
    \setlength{\tabcolsep}{3.2pt}
    \begin{tabular}{lrrllll}
    \toprule
     \multirow{2}{*}{\bf Model} &   \multirow{2}{*}{\bf Iteration}   &   \multirow{2}{*}{\bf Speed}  & \multicolumn{2}{c}{\bf En-De} &
     \multicolumn{2}{c}{\bf Ro-En} \\
    \cmidrule(lr){4-5}
    \cmidrule(lr){6-7}
    &   &   &  {\bf BLEU}&  {\bf ALF} &  {\bf BLEU} & {\bf ALF} \\
    \midrule
    \multicolumn{7}{c}{\textbf{AT Models}} \\

    {\bf Transformer-\textsc{Base}} (Ro-En Teacher) & n/a  &  1.0$\times$  & 27.3 & 70.5 &34.1 & 73.6\\
    {\bf Transformer-\textsc{Big}} (En-De Teacher)  & n/a  &  0.8$\times$  & 29.2 & 73.0 &n/a &n/a \\
    \midrule
    \multicolumn{7}{c}{\textbf{Existing NAT Models}}\\

    {\bf NAT}~\citep{NAT}    &   1.0   &   2.4$\times$ &19.2 &\multirow{5}{*}{n/a} &31.4 &\multirow{5}{*}{n/a} \\
    {\bf Iterative NAT}~\citep{lee2018deterministic} & 10.0&  2.0$\times$ & 21.6 &  & 30.2  &\\
    {\bf DisCo}~\citep{kasai2020parallel}  &   4.8 & 3.2$\times$ & 26.8 &  &33.3  &  \\
    {\bf Mask-Predict}~\citep{ghazvininejad2019mask} & 10.0  &   1.5$\times$ & 27.0  &  & 33.3 & \\
    {\bf Levenshtein}~\citep{gu2019levenshtein} &  2.5 & 3.5$\times$ & 27.3  & &33.3 & \\
    \midrule
    \multicolumn{7}{c}{\textbf{Our NAT Models}}\\
    {\bf Mask-Predict}~\citep{ghazvininejad2019mask}       &    \multirow{2}*{10.0}   &  \multirow{2}*{1.5$\times$} & 27.0 & 68.4 &33.3 & 70.9 \\
    {\bf ~~~~+Low-Frequency Rejuvenation} & &  & \bf 27.8$^\dagger$ & 72.3 & \bf 33.9$^\dagger$ & 72.4\\ 
    \hdashline
    \\[-0.95em]
    {\bf Levenshtein}~\citep{gu2019levenshtein}              &    \multirow{2}{*}{2.5}    & \multirow{2}{*}{3.5$\times$}   & 27.4  & 69.2  & 33.2 & 71.1\\
    {\bf ~~~~+Low-Frequency Rejuvenation} &  &  &\bf 28.2$^\dagger$ & 72.8 & \bf 33.8$^\dagger$ & 72.7\\
    \bottomrule
    \end{tabular}
    \caption{Comparison with previous work on  WMT14 En-De and WMT16 Ro-En. ``Iteration'' indicates the number of iterative refinement while ``Speed'' shows the speed-up ratio of decoding. ``ALF'' is the translation accuracy on low-frequency words. ``$^\dagger$'' indicates statistically significant difference ($p<0.05$) from corresponding baselines.}
    \label{tab:main-results}
\end{table*}

\section{Experiment}

\subsection{Setup}
\noindent{\bf Data}
Main experiments are conducted on four widely-used translation datasets: WMT14 English-German (En-De,~\citealt{transformer}), WMT16 Romanian-English (Ro-En,~\citealt{NAT}), WMT17 Chinese-English (Zh-En,~\citealt{hassan2018achieving}), and WAT17 Japanese-English (Ja-En,~\citealt{morishita2017ntt}), which consist of 4.5M, 0.6M, 20M, and 2M sentence pairs, respectively. We use the same validation and test datasets with previous works for fair comparison. To prove the universality of our approach, we further experiment on different data volumes, which are sampled from WMT19 En-De.\footnote{\url{http://www.statmt.org/wmt19/translation-task.html}} The {\em Small} and {\em Medium} corpora respectively consist of 1.0M and 4.5M sentence pairs, and {\em Large} one is the whole dataset which contains 36M sentence pairs.
We preprocess all data via BPE~\citep{Sennrich:BPE} with 32K merge operations. We use tokenized BLEU~\citep{papineni2002bleu} as the evaluation metric, and {\em sign-test}~\cite{collins2005clause} for statistical significance test. The translation accuracy of low-frequency words is measured by AoLC~\cite{Ding2020UnderstandingAI}, where word alignments are established based on the widely-used automatic alignment tool GIZA++~\cite{och2003systematic}.

\paragraph{Models} 
We validated our research hypotheses on two state-of-the-art NAT models:
\begin{itemize}[leftmargin=12pt]
    \item {\em Mask-Predict} (MaskT,~\citealt{ghazvininejad2019mask}) that uses the conditional mask LM~\citep{devlin2019bert} to iteratively generate the target sequence from the masked input. We followed its optimal settings to keep the iteration number as 10 and length beam as 5.
    \item {\em Levenshtein Transformer} (LevT,~\citealt{gu2019levenshtein}) that introduces three steps: {deletion}, {placeholder} and {token prediction}. The decoding iterations adaptively depends on certain conditions.
\end{itemize}
\noindent We closely followed previous works to apply sequence-level knowledge distillation to NAT~\citep{kim2016sequence}. Specifically, we train both \textsc{Base} and \textsc{Big} Transformer as the {\em AT teachers}. For \textsc{Big} model, we adopt large batch strategy (i.e. 458K tokens/batch) to optimize the performance. Most NAT tasks employ Transformer-\textsc{Big} as their strong teacher except for Ro-En and {\em Small} En-De, which are distilled by Transformer-\textsc{Base}. 

\paragraph{Training}
Traditionally, NAT models are usually trained for 300K steps on regular batch size (i.e. 128K tokens/batch). In this work, we empirically adopt large batch strategy (i.e. 480K tokens/batch) to reduce the training steps for NAT (i.e. 70K). Accordingly, the learning rate warms up to $1\times10^{-7}$ for 10K steps, and then decays for 60k steps with the cosine schedule (Ro-En models only need 4K and 21K, respectively). For regularization, we tune the dropout rate from [0.1, 0.2, 0.3] based on validation performance in each direction, and apply weight decay with
0.01 and label smoothing with $\epsilon$ = 0.1. We use Adam optimizer ~\citep{kingma2015adam} to train our models. We followed the common practices~\citep{ghazvininejad2019mask,kasai2020parallel} to evaluate the performance on an ensemble of top 5 checkpoints to avoid stochasticity. 

Note that the total training steps of the proposed approach (in \S\ref{subsec:2.2}$\sim$\ref{subsec:2.4}) are identical with those of the standard training (in \S\ref{subsec:2.1}). Taking the best training strategy ($\text{Raw}\rightarrow\vec{\text{KD}}+\cev{\text{KD}}\rightarrow\vec{\text{KD}}$) for example, we empirically set the training step for each stage is 20K, 20K and 30K, respectively. And Ro-En models respectively need 8K, 8K and 9K steps in corresponding training stage.

\begin{table}[t]
    \centering
    \begin{tabular}{lllll}
    \toprule
    \multirow{2}{*}{\textbf{Model}}   &   \multicolumn{2}{c}{\bf Zh-En}   &   \multicolumn{2}{c}{\bf Ja-En}  \\
    \cmidrule(lr){2-3} \cmidrule(lr){4-5} 
    &  {\bf BLEU} & {\bf ALF} &  {\bf BLEU} & {\bf ALF}\\
    \midrule
    {\bf\textsc{AT}} & 25.3 & 66.2 & 29.8 & 70.8\\
    \midrule
    {\bf MaskT} & 24.2 & 61.5 & 28.9 & 66.9 \\
    {\bf ~~+LFR} & 25.1$^\dagger$ & 64.8 & 29.6$^\dagger$ & 68.9\\
    \hdashline
    \\[-0.95em]
    \bf LevT & 24.4 & 62.7 & 29.1 & 66.8\\
    \bf ~~+LFR & 25.1$^\dagger$ & 65.3 & 29.7 & 69.2 \\
    \bottomrule 
    \end{tabular}
    \caption{Performance on other language pairs, including WMT17 Zh-En and WAT17 Ja-En. ``$^\dagger$'' indicates statistically significant difference ($p<0.05$) from corresponding baselines.}
    \label{tab:otherlanguage}
\end{table}

\subsection{Results}
\label{subsec:3.2}
\paragraph{Comparison with Previous Work}
Table~\ref{tab:main-results} lists the results of previous competitive NAT models~\cite{NAT,lee2018deterministic,kasai2020parallel,gu2019levenshtein,ghazvininejad2019mask} on the WMT16 Ro-En and WMT14 En-De benchmark. We implemented our approach on top of two advanced NAT models (i.e. Mask-Predict and Levenshtein Transformer).
Compared with standard NAT models, our training strategy significantly and consistently improves translation performance (BLEU$\uparrow$) across different language pairs and NAT models. Besides, the improvements on translation performance are mainly due to a increase of translation accuracy on low-frequency words (ALF$\uparrow$), which reconfirms our claims. For instance, our method significantly improves the standard Mask-Predict model by +0.8 BLEU score with a substantial +3.6 increase in ALF score. Encouragingly, our approach push the existing NAT models to achieve new SOTA performances (i.e. 28.2 and 33.9 BLEU on En-De and Ro-En, respectively).

It is worth noting that our data-level approaches neither modify model architecture nor add extra training loss, thus do not increase any latency (``Speed''), maintaining the intrinsic advantages of non-autoregressive generation. We must admit that our strategy indeed increase the amount of computing resources due to that we should train $f_{t \mapsto s}$ AT teachers for building $\cev{\text{KD}}$ data.

\paragraph{Results on Other Language Pairs}

Table~\ref{tab:otherlanguage} lists the results of NAT models on Zh-En and Ja-En language pairs, which belong to different language families (i.e. Indo-European, Sino-Tibetan and Japonic). Compared with baselines, our method significantly and incrementally improves the translation quality in all cases.
For Zh-En, LFR achieves on average +0.8 BLEU improvement over the traditional training, along with increasing on average +3.0\% accuracy on low-frequency word translation. For long-distance language pair Ja-En, our method still improves the NAT model by on average +0.7 BLEU point with on average +2.2 ALF.  Furthermore, NAT models with the proposed training strategy perform closely to their AT teachers (i.e. 0.2 $\Delta$BLEU).
This shows the effectiveness and universality of our method across language pairs.

\begin{table}[t]
\centering
\begin{tabular}{llllll}
\toprule
	\bf Model   &  \bf Law & \bf Med. & \bf IT & \bf Kor. & \bf Sub. \\
	\midrule
	\bf AT &41.5 &30.8 &27.5 &8.6 &15.4 \\
    \midrule
    \bf MaskT   &37.3 &28.2 &24.6 &7.3 &11.2\\
    \bf ~~+LFR  &38.1$^\dagger$ &28.8 &25.4$^\dagger$ &8.9$^\dagger$ &14.3$^\dagger$ \\
    \hdashline
    \\[-0.95em]
    \bf LevT & 37.5 & 28.4 & 24.7 & 7.5 & 12.4 \\
    \bf ~~+LFR & 38.5$^\dagger$ & 29.4$^\dagger$ & 25.9$^\dagger$ & 8.4$^\dagger$ & 14.5$^\dagger$ \\
    \bottomrule
\end{tabular}
\caption{Performance on domain shift setting. Models are trained on WMT14 En-De news domain but evaluated on out-of-domain test sets, including law, medicine, IT, koran and subtitle. ``$^\dagger$'' indicates statistically significant difference ($p<0.05$) from corresponding baselines.}
\label{tab:domain-shift}
\end{table}

\paragraph{Results on Domain Shift Scenario}
The lexical choice must be informed by linguistic knowledge of how the translation model's input data maps onto words in the target domain. 
Since low-frequency words get lost in traditional NAT models, the problem of lexical choice is more severe under domain shift scenario (i.e. models are trained on one domain but tested on other domains). Thus, we conduct evaluation on WMT14 En-De models over five out-of-domain test sets~\cite{Muller:2019tq}, including law, medicine, IT, Koran and movie subtitle domains. 
As shown in Table \ref{tab:domain-shift}, standard NAT models suffer large performance drops in terms of BLEU score (i.e. on average -2.9 BLEU over AT model). By observing these outputs, we found a large amount of translation errors on low-frequency words, most of which are domain-specific terminologies. In contrast, our approach improves translation quality (i.e. on average -1.4 BLEU over AT model) by rejuvenating low-frequency words to a certain extent, showing that LFR increases the domain robustness of NAT models. 

\paragraph{Results on Different Data Scales}

\begin{table}
\centering
    \begin{tabular}{llll}
    \toprule
    \multirow{2}{*}{\textbf{Model}}   &   \multicolumn{3}{c}{\bf BLEU} \\
    \cmidrule(lr){2-4}
    &   \bf 1.0M & \bf 4.5M  & \bf 36.0M \\
    \midrule
    {\bf\textsc{AT}} & 25.5 &37.6 &40.2 \\
    \midrule
    {\bf MaskT}  & 23.7 &35.4 &36.8 \\
    \bf ~~+LFR  & 24.3$^\dagger$ & 36.2$^\dagger$ & 37.7$^\dagger$ \\
    \bottomrule
    \end{tabular}
    \caption{Performance on different scale of training data. The small and medium datasets are sampled from the large WMT19 En-De dataset, and evaluations are conducted on the same testset. ``$^\dagger$'' indicates statistically significant difference ($p<0.05$) from corresponding baselines.}
    \label{tab:scale-result}
\end{table}

To confirm the effectiveness of our method across different data sizes, we further experiment on three En-De datasets at different scale. The small- and medium-scale training data are randomly sampled from WM19 En-De corpus, containing about 1.0M  and 4.5M sentence pairs, respectively. The large-scale one is collected from WMT19, which consists of 36M sentence pairs. We report the BLEU scores on same testset \texttt{newstest2019} for fair comparison. We employs base model to train the small-scale AT teacher, and big model with large batch strategy (i.e. 458K tokens/batch) to build the AT teachers for  medium- and large-scale.
As seen in Table~\ref{tab:scale-result}, our simple training recipe boost performances for NAT models across different size of datasets, especially on large scale (+0.9), showing the robustness and effectiveness of our approach.

\paragraph{Complementary to Related Work}
\citet{Ding2020UnderstandingAI} is relevant to our work, which introduced an extra bilingual data-dependent prior objective to augment NAT models the ability to learn low-frequency words in raw data. Our method is complementary to theirs due to that we only change data and training strategies (model-agnostic). As shown in Table~\ref{tab:complementary},
two approaches yield comparable performance in terms of BLEU and ALF. Besides, combination can further improve BLEU as well as ALF scores (i.e. +0.3 and +0.6). This illustrates the complementarity of model-level and data-level approaches on rejuvenating low-frequency knowldege for NAT models.

\subsection{Analysis}
\label{sec:analysis}
We conducted extensive analyses to better understand our approach. All results are reported on the Mask-Predict models.

\paragraph{Accuracy of Lexical Choice}
To understand where the performance gains come from, we conduct fine-grained analysis on lexical choice. We divide ``All'' tokens into three categories based on their frequency, including ``High'', ``Medium'' and ``Low''. 
Following \citet{Ding2020UnderstandingAI}, we measure the accuracy of lexical choice on different frequency of words. Table~\ref{tab:lexical} shows the results. 
{\bf Takeaway:}
\textit{The majority of improvements on translation accuracy is from the low-frequency words, confirming our hypothesis.}

 \begin{table}[t]
    \centering
    \begin{tabular}{llll}
    \toprule
    \bf Model & \bf BLEU & \bf ALF\\
    \midrule
    \bf Mask-Predict & 27.0 & 68.4 \\
    \bf ~~+Raw Data Prior & 27.8 & 72.4\\
    \bf ~~+Low-Frequency & 27.8 & 72.3\\
    \midrule
    \bf ~~+Combination & 28.1 &72.9\\
    \bottomrule
    \end{tabular}
\caption{Complementary to other work. ``Combination'' indicates combining ``+Raw Data Prior'' proposed by \citet{Ding2020UnderstandingAI} with our ``+Low-Frequency''. Experiments are conducted on WMT14 En-De.}
\label{tab:complementary}
\end{table}

\begin{table*}[t]
\centering
\setlength{\tabcolsep}{4.2pt}
    \begin{tabular}{lcccccccccccc}
    \toprule
    \multirow{2}{*}{\textbf{Model}}   &   \multicolumn{4}{c}{\bf En-De} & \multicolumn{4}{c}{\bf Zh-En} & \multicolumn{4}{c}{\bf Ja-En}\\
    \cmidrule(lr){2-5} \cmidrule(lr){6-9} \cmidrule(lr){10-13}
    & All & High & Med. & Low & All & High & Med. & Low & All & High & Med. & Low \\
    \midrule
    \bf MaskT (Raw) &74.3 &75.9 &74.6 &72.5 &68.5 &71.5 &68.3 &65.1 &73.1 & 75.5 &74.7 & 69.1 \\
    \bf MaskT (KD) &76.3 &82.4 &78.3 &68.4 &72.7 &81.4 &75.2 &61.5 &75.3 &82.8 &76.3 &66.9\\
    \midrule
    {\bf +Raw-Pretrain}  & \cellcolor{b2} 77.7 & \cellcolor{b1} 83.1 & 78.4 & \cellcolor{b4} 71.9 & \cellcolor{b1} 73.4 & 81.6 & 75.3 & \cellcolor{b3} 64.1 & \cellcolor{b1} 76.1 & \cellcolor{b1} 83.4 & \cellcolor{b1} 76.7 & \cellcolor{b2} 68.3\\
    {\bf ~~+Bi-Distillation} & \cellcolor{b2} 77.9 & \cellcolor{b1} 83.1 & 78.5 & \cellcolor{b4} 72.3 & \cellcolor{b2} 73.7 & 81.7 & 75.3 & \cellcolor{b4} 64.8 & \cellcolor{b2} 76.5 & \cellcolor{b1} 83.5 & \cellcolor{b1} 76.7 & \cellcolor{b3} 68.9\\
    \bottomrule
    \end{tabular}
    \caption{Analysis on different frequency words in terms of accuracy of lexical choice. We split ``All'' words into ``High'', ``Medium'' and ``Low'' categories. 
    Shades of cell color represent differences between ours and KD.}
    \label{tab:lexical}
\end{table*}

\paragraph{Low-Frequency Words in Output}
We expect to recall more low-frequency words in translation output. 
As shown in Table~\ref{tab:target-frequency}, we calculate the ratio of low-frequency words in generated sentences. As seen, KD biases the NAT model towards generating high-frequency tokens (\textit{Low freq.}$\downarrow$) while our method can not only correct this bias (on average +18\% and +26\% relative changes for \textit{+raw-pretrain} and \textit{+Bi-distillation}), but also enhance translation (BLEU$\uparrow$ in Table~\ref{tab:otherlanguage}). 
{\bf Takeaway:} \textit{Our method generates translations that contain more low-frequency words.} 

\begin{table}
\centering
\begin{tabular}{lrrr}
\toprule
\bf Model   &   \bf En-De   &   \bf Zh-En   &   \bf Ja-En \\
\midrule
\bf MaskT (Raw)             &   10.3\% & 6.7\% & 9.4\% \\
\bf MaskT (KD)    &   7.6\%  & 4.2\% & 6.9\% \\
\midrule
\bf ~~{+Raw-Pretrain}         &9.3\%  &5.6\% &8.4\% \\
\bf ~~~~{+Bi-Distillation}         &{\bf 9.7}\% &{\bf 6.8}\% &{\bf 8.7}\% \\
\bottomrule
\end{tabular}
\caption{Ratio of low-frequency target words in output.} 
\label{tab:target-frequency}
\end{table}

\begin{table}[t]
    \centering
    \begin{tabular}{clll}
    \toprule
    \bf \# & \bf Strategy & \bf BLEU & \bf ALF\\
    \midrule
    1 & Raw &24.1 &69.3 \\
    2 & $\vec{\text{KD}}$ &25.4 &66.4 \\
    \midrule
    3 & Raw+$\vec{\text{KD}}$ &25.6 &67.7 \\
    4 & Raw$\rightarrow$$\vec{\text{KD}}$ &\bf 25.9 &68.2 \\
    \hdashline
    \\[-0.95em]
    5 & Raw+$\cev{\text{KD}}$+$\vec{\text{KD}}$ &25.7 &67.9 \\
    6 & Raw$\rightarrow$$\cev{\text{KD}}$+$\vec{\text{KD}}$ &25.7 &68.3 \\
    7 & Raw$\rightarrow$$\cev{\text{KD}}$+$\vec{\text{KD}}$$\rightarrow$$\vec{\text{KD}}$ &\bf 26.3 &69.5 \\
    \bottomrule
    \end{tabular}
\caption{Performances of different strategies. The models are trained and tested on WMT14 En-De. ``A+B'' means concatenate A and B while ``A$\rightarrow$B'' indicates pretraining on A and then finetuning on B.}
\label{tab:variant-strategy}
\end{table}

\paragraph{Effects of Variant Training Strategies}
As discussed in \S\ref{subsec:2.4}, we carefully investigate alternative training approaches in Table~\ref{tab:variant-strategy}. We make the total training step identical to that of vanilla NAT models, and report both BLEU and ALF scores. 
As seen, all variant strategies perform better than the standard KD method in terms both BLEU and ALF scores, confirming the necessity of our work.
{\bf Takeaway:} \textit{1) Pretraining is more effective than combination on utilizing data manipulation strategies; 2) raw data and bidirectional distilled data are complementary to each other; 3) it is indispensable to finetune models on $\vec{\text{KD}}$ in the last stage.}

\begin{table}[t]
\centering 
    \begin{tabular}{lcccc}
    \toprule
    \multirow{1}{*}{\textbf{Model}} 
    & All & High & Med. & Low \\
    \midrule
    \multicolumn{5}{c}{\em Training on Raw Data}\\
    \bf AT-Teacher &79.3 &84.7 &80.2 &73.0 \\
    \bf AT-Student &76.8 &80.2 &77.4 &72.8 \\ 
    \midrule
    \multicolumn{5}{c}{\em Training on Distilled Data}\\
    \bf AT-Student &77.3 &82.5 &78.6 &70.9 \\ 
    \bf ~~+LFT &78.1 &83.2 &78.7 & 72.5\\
    \bottomrule
    \end{tabular}
    \caption{Analysis on AT models in term of the accuracy of lexical choice on WMT14 En-De. We split ``All'' words into ``High'', ``Medium'' and ``Low'' categories.}
    \label{tab:at-models}
\end{table}

\paragraph{Our Approach Works for AT Models}
Although our work is designed for NAT models, we also investigated whether our LFT method works for general cases, e.g. autoregressive models. We used Transformer-\textsc{Big} as the teacher model. For fair comparison, we leverage the Transformer-\textsc{Base} as the student model, which shares the same model capacity with NAT student (i.e. MaskT). The result lists in Table~\ref{tab:at-models}. As seen, AT models also suffer from the problem of low-frequency words when using knowledge distillation, and our approach also works for them. {\bf Takeaway:} \textit {Our method works well for general cases through rejuvenating more low-frequency words.}

\section{Related Work}

\paragraph{Low-Frequency Words}

Benefiting from continuous representation learned from the training data, NMT models have shown the promising performance. 
However, \newcite{koehn-knowles-2017-six} point that low-frequency words translation is still one of the key challenges for NMT according to the Zipf's law~\cite{zipf1949human}. 
For AT models,~\citet{arthur2016incorporating} address this problem by integrating a count-based lexicon, and~\citet{nguyen-chiang-2018-improving} propose an additional lexical model, which is jointly trained with the AT model. Recently, \newcite{gu2020token} adaptively re-weight the rare words during training.
The lexical choice problem is more serious for NAT models, since 1) the lexical choice errors (low-resource words in particular) of AT distillation will propagate to NAT models; and 2) NAT lacks target-side dependencies thus misses necessary target-side context. In this work, we alleviate this problem by solving the first challenge. 

\paragraph{Data Manipulation}
Our work is related to previous studies on manipulating training data for NMT. \newcite{Bogoychev2019DomainTA} show that forward- and backward-translations (FT/ BT) could both boost the model performances, where FT plays the role of domain adaptation and BT makes the translation fluent. \newcite{fadaee-monz-2018-back} sample the monolingual data with more difficult words (e.g. rare words) to perform BT, achieving significant improvements 
compared with randomly sampled BT. \newcite{nguyen2020data} diversify the data by applying FT and BT multiply times. However, different from AT, the prerequisite of training a well-performed NAT model is to perform KD. We compared with related works in Table~\ref{tab:variant-strategy} and found that our approach consistently outperforms them. Note that all the ablation studies focus on exploiting the parallel data without augmenting additional data.

\paragraph{Non-Autoregressive Translation}

A variety of approaches have been exploited to bridge the performance gap between NAT and AT models. Some researchers proposed new model architectures~\citep{lee2018deterministic, ghazvininejad2019mask, gu2019levenshtein, kasai2020parallel}, aided with additional signals~\citep{wang2019non,Ran2019GuidingNN,ding-etal-2020-context}, introduced sequential information~\citep{wei-etal-2019-imitation,shao2019minimizing,guo2020fine,hao-etal-2021-multi}, and explored advanced training objectives~\citep{ghazvininejad2020aligned,Du:2021:ICML}. Our work is close to the research line on training methods. ~\citet{Ding2020UnderstandingAI} revealed the low-frequency word problem in distilled training data, and introduced an extra Kullback-Leibler divergence term derived by comparing the lexical choice of NAT model and that embedded in the raw data. \citet{ding2021progressive} propose a simple and effective training strategy, which progressively feeds different granularity of data into NAT models by leveraging curriculum learning. 

\section{Conclusion}

In this study, we propose simple and effective training strategies to rejuvenate the low-frequency information in the raw data.
Experiments show that our approach consistently and significantly improves translation performance across language pairs and model architectures.
Notably, domain shift is an extreme scenario to diagnose low-frequency translation, and our method significant improves them. Extensive analyses reveal that our method improves the accuracy of lexical choices for low-frequency source words, recalling more low-frequency words in translations as well, which confirms our claim. 

\section*{Acknowledgments}
We are grateful to the anonymous reviewers and the area chair for their insightful comments and suggestions. Xuebo Liu and Derek F. Wong were supported in part by the Science and Technology Development Fund, Macau SAR (Grant No. 0101/2019/A2), and the Multi-year Research Grant from the University of Macau (Grant No. MYRG2020-00054-FST).

\bibliographystyle{acl_natbib}
\bibliography{acl2021}

\newpage
\appendix
\end{document}